\documentclass{svproc}
 \usepackage{cuted} 
\usepackage{cite}
\usepackage{amsmath,amssymb,amsfonts,nccmath}
\usepackage{cleveref}
\usepackage{algorithmic}
\usepackage[strings]{underscore}
\usepackage{float}

\usepackage{graphicx}
\usepackage{textcomp}
\usepackage{xcolor}

\usepackage{url}

\begin{document}
\mainmatter              
%
\title{Aztec curve: proposal for a new space-filling curve\\}
\titlerunning{Aztec curve}  
%
\author{Ayala Diego\inst{1} \ \inst{0000-0003-3063-2714} \and Durini Daniel\inst{1} \ \inst{0000-0002-5979-1391} \and Rangel-Magdaleno Jose\inst{1} \ \inst{0000-0003-2785-5060}}
\authorrunning{Ayala Diego et al.} 
%
\tocauthor{Ayala Diego, Durini Daniel, Rangel-Magdaleno Jose}
\institute{$ ^{1} $INAOE, Puebla, Mexico\\
\email{diegoayala002@gmail.com}, \email{ddurini@inaoep.mx}, \email{jrangel@inaoep.mx}}
\maketitle      

\begin{abstract}
	Different space-filling curves (SFCs) are briefly reviewed in this paper, and a new one is proposed. A century has passed between the inception of this kind of curves, since then they have been found useful in computer science, particularly in data storage and indexing due to their clustering properties, being Hilbert curve the most well-known member of the family of fractals. The proposed Aztec curve, with similar characteristics to the Hilbert's curve, is introduced in this paper, accompanied by a grammatical description for its construction. It yields the possibility of creating bi-dimensional clusters, not available for Hilbert nor Peano curves. Additional to this, a case of application on the scope of Compressed Sensing is implemented, in which the use of Hilbert curve is contrasted with Aztec curve, having a similar performance, and positioning the Aztec curve as viable and a new alternative for future exploitation on applications that make use of SFC's.

\end{abstract}
\section{Introduction}

\begin{figure}[htbp]
	\centerline{\includegraphics[width=0.8\textwidth]{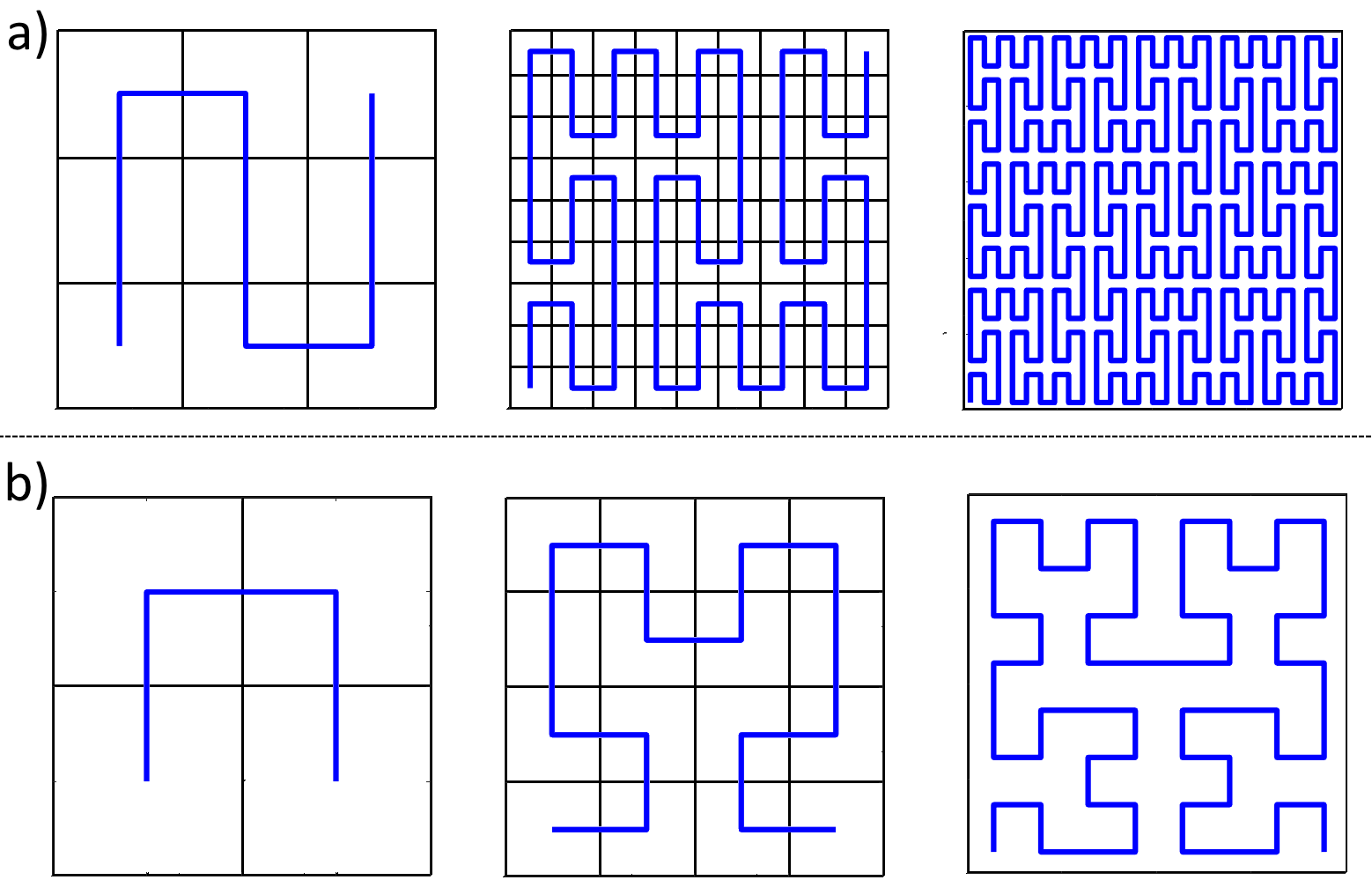}}
	\caption[Peano and Hilbert curves.] {The first, second, and third order of: a) Peano curve and b) Hilbert curve.}
	\label{fig:peano}	
\end{figure}

It is considered that the first space-filling curve (SFC) was presented in 1890 by G. Peano, who described a continuous curve that goes through every coordinate point of the unit square \cite{peano}. The following year D. Hilbert proposed his own curve \cite{Hilbert}. Both curves pass through the center points of all inner squares that are enclosed by the outer  square unit, and exhibit self-similar patterns at smaller scales, being each scale referred as an 'order', and represented as $n\in\mathbb{R}^{+}$ . As there is no upper limit for $ n $, the patterns can be replicated an infinite number of times without exceeding the boundary of the square unit, as illustrated in Fig. \ref{fig:peano}.

Initially, space-filling curves were considered no more than mathematical curiosities, being even called "mathematical monsters" by Goldschager in 1981 \cite{Bader2013}, around the time a variety of curves had been proposed. Great attention was drawn to these mathematical artifacts due to their intrincate and elegant recursive structure. They belong to the family of fractal curves discussed by B. Mandelbrot \cite{mandelbrot1983}, who coined and popularized the term "fractal".

Later, on 1988 the so-called Generic Space-Filling Heuristic (GSFH) was formulated by Batholdi and Platzman \cite{Bartholdi1988}, which consisted on using SFCs to translate a combinatorial problem from the Euclidean space to the unit interval, which yields an easier solution. The authors applied this heuristic to solve the Travelling Salesman Problem (TSP), which is well-known on the realm of Computer Sciences, exhibiting solutions close to the optimal. Since then, SFC have been proven to have important locality properties which are useful to arrange and access neighboring elements inside data structures \cite{Bader2013}. The following examples illustrate some direct applications of the SFCs relevant to date: improving the quality partitioning in parallel computing \cite{parallel}, defining the "S2Cells" path (which is a key library used by GoogleMaps) based on the Hilbert curve \cite{s2} to identify the regions ranging from millimeters to kilometers on Earth's surface, designing more efficient planar resonant antennas using the physical path of the Hilbert curve \cite{Vinoy2001}, or developing a set of Lumped Element Kinetic Inductance Detectors (LEKID) used inside the NIKA2 instrument of the 30 m IRAM telescope located at Pico Valeta in Spain \cite{nika2}.
\begin{figure}[htbp]
	\centerline{\includegraphics[width=0.6\textwidth]{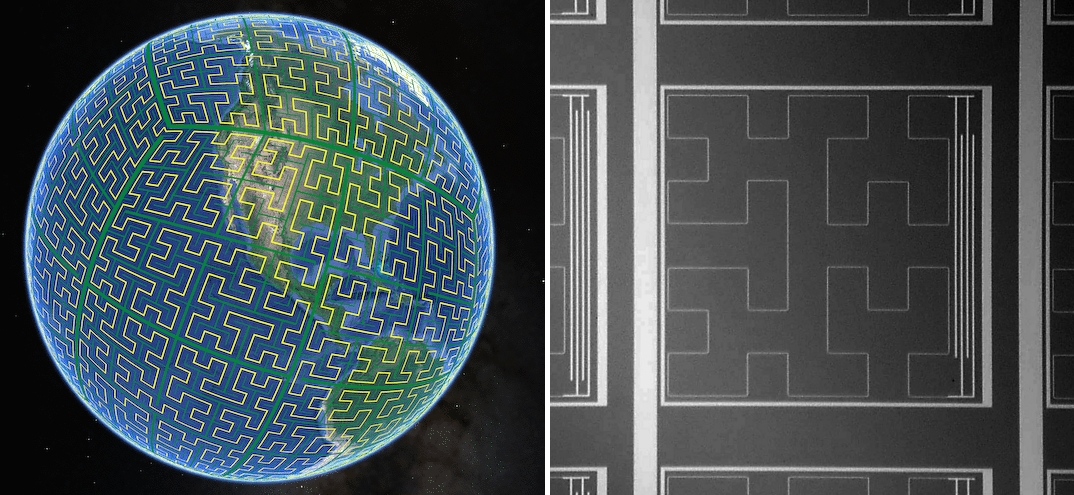}}
	\caption[Applications of the Hilbert curve.] {Applications of the Hilbert curve on: a) S2Cells projected onto the sphere \cite{s2}, b) a front-illuminated microstrip pixel for the 260 GHz on NIKA2 \cite{nika2}.}
	\label{fig:hilbertapp}		
\end{figure}

Not all SFC's tend to cross have a correspondence to adjacent sub-squares, as it is the case, for example, of the so-called Lebesgue's curve \cite{Bader2013}, nor are they restrained to square-like figures, as it is shown in the case of the Gosper curve that fills hexagonal shape figures instead \cite{Bader2013}. Moreover, different SFC paths, as it is shown in Fig. \ref{fig:peano}, do not necessarily divide the square unit following the same rate of sub-divisions. For example, the Peano curve is based on iterations that cover $9^{n}$ squares, whilst the Hilbert covers $4^{n}$ squares.

\section{Aztec curve proposal}
\begin{figure}[htbp]
	\centerline{\includegraphics[width=0.6\textwidth]{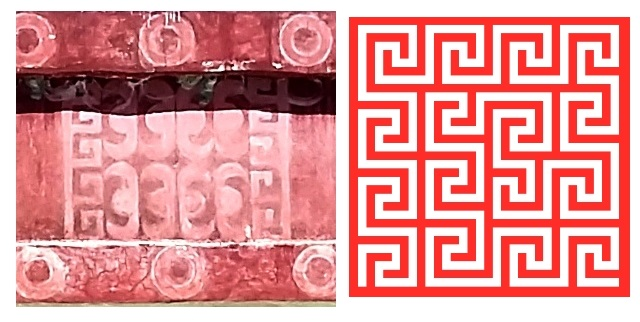}}
	\caption[Teotihuacan palace and first order Aztec curve.] {a) Detail on Quetzalpapálotl Palace. b) Second order Aztec curve.}
	\label{fig:aztecfirst}		
\end{figure}

In this work, we propose another curve that we named the "Aztec curve", as it was partially inspired by the ornaments discovered on the Quetzalpapálotl Palace located in the archaelogical site of Teotihuacán in Mexico (see Fig. \ref{fig:aztecfirst}).\\

This curve is a unit interval $ [0,1] $ that has a one-to-one mapping onto the square unit $ [0,1]^{2} $, and is defined such as: the square unit is subdivided into sixteen sub-squares, each of them with sides quarter of the length of the father square. A curve fill each sub-square via rotations, or reflections of itself. The sixteen curves are connected through the operators of rotation, and reflection in a way that preserve its continuity thorough all the sub-squares.The curve is chosen along with the operators in order to fill the defined square unit,  grows into self similar segments at smaller scales, and if iterated, resembles a meander-shape figure.

\subsection{Grammar-based description}
\begin{figure}[htbp]
	\centerline{\includegraphics[width=0.6\textwidth]{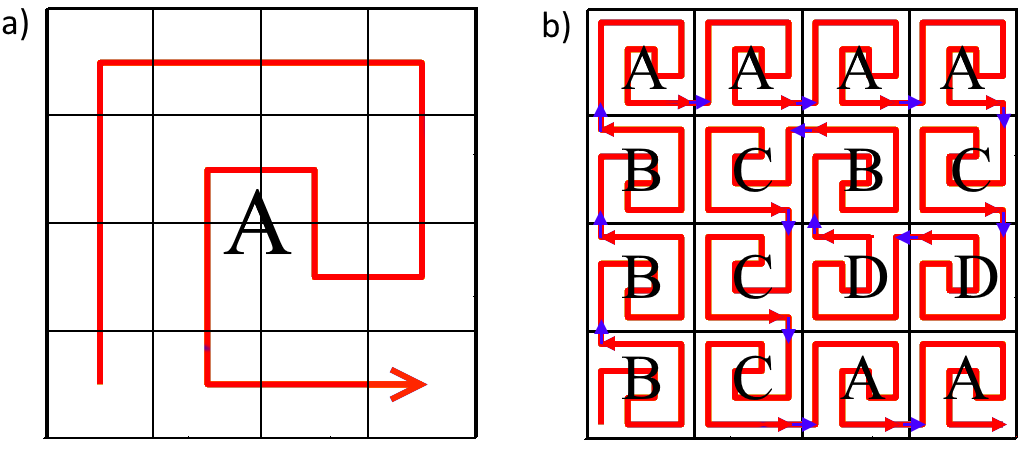}}
	\caption[a) First-order Aztec curve, b) Second-order Aztec curve is constructed by rotations and flips of its predecessor.] {a) First-order Aztec curve, b) Second-order Aztec curve is constructed by rotations and flips of it's predecessor.}
	\label{fig:aztec2}		
\end{figure}

To describe the iterations of the proposed Aztec curve, the grammar description defined by M. Bader \cite{Bader2013} will be used. The unit square is divided into 16 sub-squares which will be populated by line segments that follow the path of the first-order curve $ A_1$, which is the original pattern (see Fig.\ref{fig:aztec2}a). A set of non-terminal symbols \textit{\{A,B,C,D\}} represents the following: \textit{A} the n-order path, \textit{B} a reflection and 90º clockwise rotation of \textit{A}, \textit{C} is a 180º rotation of \textit{B}, and \textit{D} is a a 180º rotation of \textit{A}. The terminal symbols \{$\uparrow$,	$\downarrow$,	$\leftarrow$,$\rightarrow$\} describe the translations between the sub-squares. A grammatical description of how to construct an n-order Aztec curve (for $ n=k+1 $), is given as:
\begin{fleqn}
	\begin{multline}
		A_{k+1}\Leftarrow B_{k}\uparrow B_{k}\uparrow B_{k}\uparrow A_{k}\rightarrow A_{k}\rightarrow A_{k}\rightarrow A_{k} \downarrow C_{k} \downarrow D_{k}\leftarrow D_{k}\uparrow B_{k} 
		\leftarrow C_{k} \\ \downarrow C_{k}\downarrow C_{k}\rightarrow A_{k}\rightarrow A _{k} \label{eqn:an}
	\end{multline}
\end{fleqn}
\begin{equation}
	B_{k}=R_{O,-90º}(refl(A_{k}))	\label{eqn:bn}
\end{equation}
\begin{equation}
	C_{k}=R_{O,180º}(B_{k})	\label{eqn:cn}
\end{equation}
\begin{equation}
	D_{k}=R_{O,180º}(A_{k})	\label{eqn:dn}
\end{equation}
Equation \eqref{eqn:an} for $ k=1 $, describes the second-order curve $ A_2 $, which is equivalent to a sequence of first-order non-terminal symbols (defined in \eqref{eqn:bn}, \eqref{eqn:cn}, \eqref{eqn:dn}) placed on a path by the terminal symbols (arrows); on Fig. \ref{fig:aztec2} the above-mentioned proceeding is visually represented.

\begin{figure}[h]
	\centerline{\includegraphics[width=0.55\textwidth]{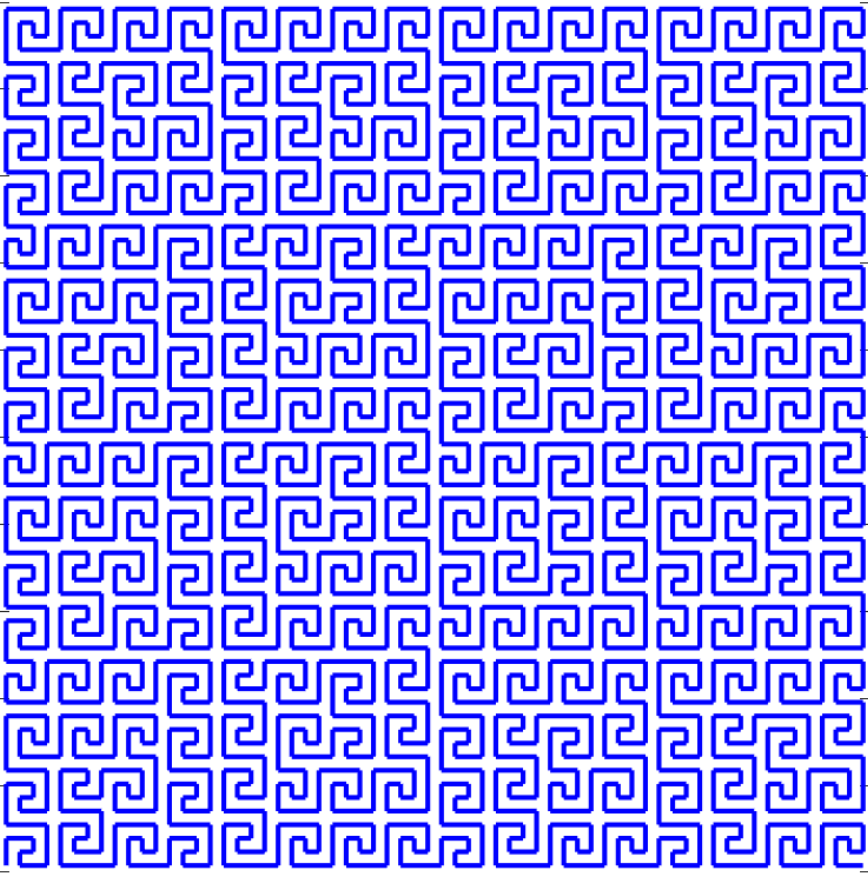}}
	\caption[Third-order Aztec curve.] {Third-order Aztec curve.}
	\label{fig:aztec3}		
\end{figure}

The next iteration would be a third-order Aztec curve (depicted on Fig. \ref{fig:aztec3}), which is thus constituted by a sequence of second-order replacements of \textit{\{A2,B2,C2,D2\}} spatially arranged by the same terminal symbols shown in \eqref{eqn:an}. This procedure can be repeated infinite times, producing self-similar patterns at smaller scales, a property known from fractals.

\subsection{Clustering properties}
\begin{figure}[h]
	\centerline{\includegraphics[width=0.65\textwidth]{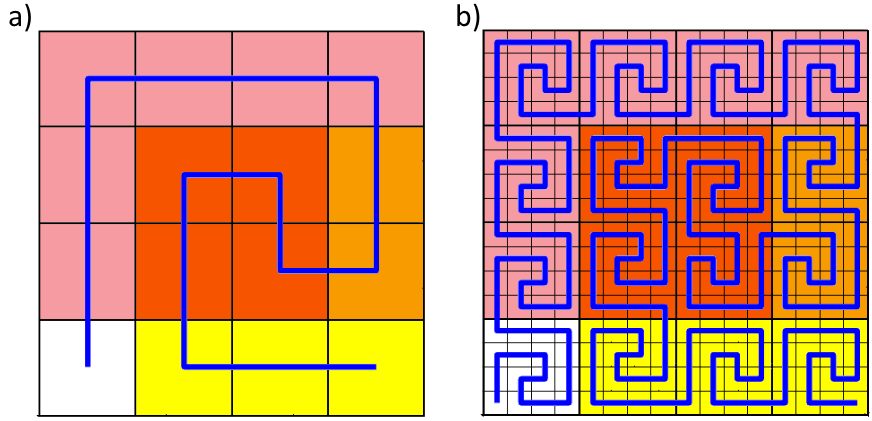}}
	\caption[Clusters.] {Rectangular subsets covering a continuous path of $4^{2n-1}$ (dark orange), $9\times16^{n-1}$ (yellow), $2\times3\times16^{n-1}$ (light orange), and $3\times4\times16^{n-1}$ (pink) sub-squares, are overlayed on: a) first-order Aztec curve, a) second-order Aztec curve.}
	\label{fig:clusters}		
\end{figure}

Although the number of sub-squares filled by the Aztec curve equals $4^{2n}$, it can contain various rectangular sub-sets smaller in size, each one housing a continuous path that crosses all the adjacent sub-squares, this will be referred to as clusters.\\
If each sub-square is considered to be a memory cell, then it becomes apparent that the Aztec curve can be used to store clusters of $4^{2n}$ size, moreover its structure allows the existence of internal rectangular clusters that, under other paths, would rather be non-connected. To illustrate this subsets with different proportions are shown in Fig. \ref{fig:clusters}.\

Therefore the Aztec curve has a correspondence in the 2D-space that harbors sets of clusters consisting of: $4^{2n}$ sub-squares, $4^{2n-1}$ sub-squares (which in conjunction with the $4^{2n}$ conform a set that have a one-to-one correspondence to the set available to Hilbert curve), and $9\times16^{n-1}$ sub-squares
Hence, sets of clusters that are not-continuous for Hilbert curve, nor Peano curve, are contained within the Aztec curve as a continuous path, which makes it suitable to use as an alternative to index and access data as well as other applications.

\section{Case of application}
A case of application of the Aztec curve is proposed on the scope of Compressed Sensing (CS), which is an advanced signal processing technique proposed in 2006 by Donoho \cite{Donoho}, and whose principle is that a signal sparse in nature can be efficiently reconstructed by a number of measurements much smaller than the length of the digitally sampled signal \cite{Stankovic}. Likewise a digital signal can be  compressed and reconstructed from a set of sparse non-zero coefficients which are computed through a Matching Pursuit (MP) algorithm, proposed by S. Mallat et al. \cite{Mallat1993MatchingPW}, decomposing the signal to a linear combination of coefficients with a series of wavelets from a fixed dictionary, selecting waveforms that has the best fit compared to the original signal, and calculating the weight of this waveforms as a few sparse coefficients. If the MP algorithm is considered as a black-box it can process a 2D image through the use of a SFC converting it to a 1D signal, the original image can be reconstructed through a linear combination of the output coefficients with the known dictionary.\\

\begin{figure}[h]
	\centerline{\includegraphics[width=0.85\textwidth]{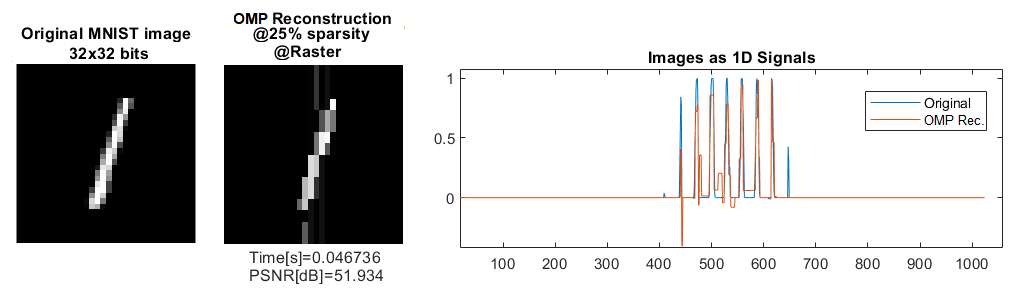}}
	\caption[An image from the MNIST database, processed through the OMP algorithm.] {An image from the MNIST database, processed through the OMP algorithm.}
	\label{fig:ompex}		
\end{figure}

A test is conducted over the publicly available MNIST database \cite{mnist}, which consist of 28-by-28 pixel images of white handwritten numbers on a black background. The image on Fig. \ref{fig:ompex} is padded with black pixels until it is 32-by-32 in size, then it is arranged as a 1D vector column-by-column (thorough a vertical raster scan), and fed to the stock MATLAB's Orthogonal Matching Pursuit (OMP) algorithm, which was proven as a slight improve over MP \cite{omp}, afterwards it calculates the sparse coefficients based on the Haar wavelet dictionary, and finally the image is reconstructed as the linear combination of the sparse coefficients with the corresponding wavelets. 

\begin{figure}[h]
	\centerline{\includegraphics[width=0.95\textwidth]{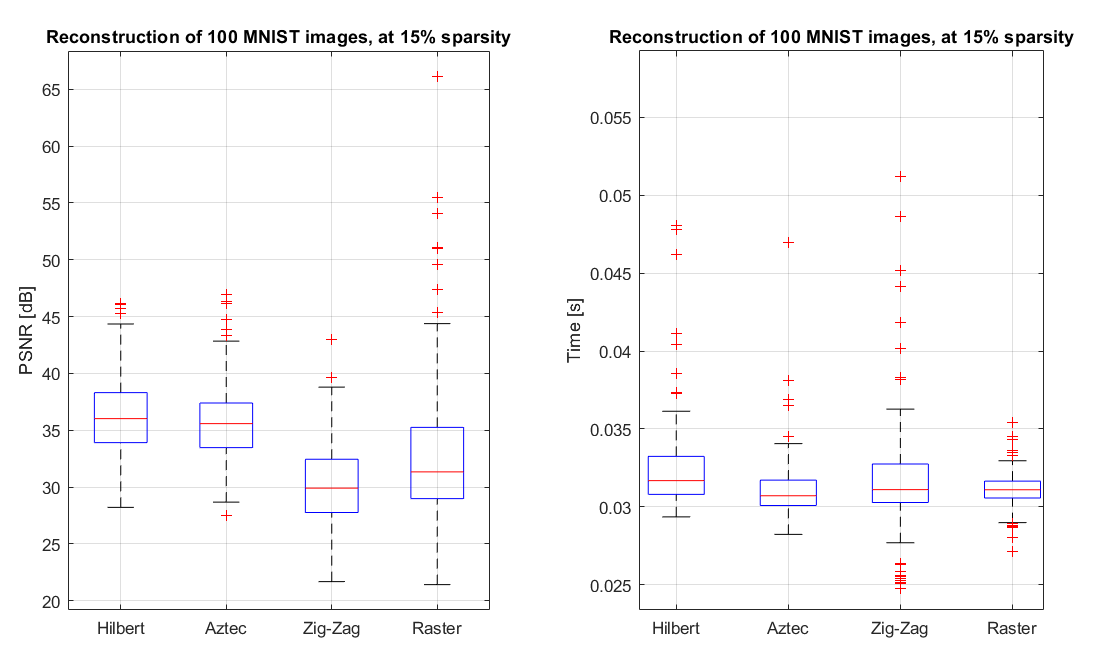}}
		\caption[Box-plots from batches of 100 MNIST database images reconstructed by OMP algorithm under different scanning methods; each batch was fed through different scanning methods.] {Box-plots from batches of 100 MNIST database images reconstructed by OMP algorithm under different scanning methods; each image from the batch was fed through the following methods: Hilbert curve, Aztec curve, Zig-Zag scan, and Raster scan.}
	\label{fig:boxplot}		
\end{figure}

The sparsity value is often an input to MP algorithms and limits the number of non-zero coefficients which are only a fraction of the total length of the signal, consequently an image is not reconstructed as a copy of the original, and it can be considered as a lossy compression, interestingly enough it can be observed on Fig. \ref{fig:ompex} that some vertical artifacts are introduced.
The padding of black pixels permits this images to be 1024 pixels in size, which enables it to be rearranged as a one-dimensional vector by either Hilbert, or Aztec curves, which in turn may cause an impact on the performance of the OMP algorithm. To asset this, a batch of 100 images is subject to a similar test, but the rearrangement of the one-dimensional vector is made through 4 different methods: Hilbert curve, Aztec curve, Zig-Zag scan, and Raster scan. To provide an statistical insight the processing time and the Peak signal-to-noise ratio of the batches are represented as box-plots in Fig. \ref{fig:boxplot}.

From the box-plots of Fig. \ref{fig:boxplot} it can be seen that both SFC's yield a better PSNR than Zig-Zag, and raster scans, being the median of the batch rearranged by Hilbert curve meagerly greater than that of the Aztec curve; on the other hand there seems to be an slight reduction in processing time when the Aztec curve is used, exhibiting a lower median, and overall more consistent times when compared to Hilbert curve. In consequence the Aztec curve is not only an alternative to the Hilbert curve, but closely match the advantages on performance, in this particular application of Compressed Sensing.

\section{Conclusions}
A new SFC called Aztec curve was presented that is designed to fill the defined square-units and remain self-similar at reduced scales of observation, as do also the well-known Hilbert and Peano curves. The Aztec curve may offer a new set of possibilities that can outperform similar space-filling curves if applied to different applications, such as, for example, those used in Compressed Sensing.
The clustering properties of the Aztec curve allow for connection of sets of adjacent sub-squares in a continuous uninterrupted manner, which might be used in efficient data storage or accesing applications, especially in two-dimensional arrays.

\section*{Acknowledgments}

D. A. acknowledges INAOE and its staff, for their academic excellence even on the face of the current world pandemic, and under the support of CONACYT's scholarship (CVU:1016578). Also special thanks to Alexandra Elbakyan and Aaron Swartz, for making research accessible to any person in this world, regardless of place of birth, economic level, or affiliation.

%
%
\bibliographystyle{spmpsciunsrt.bst}
\bibliography{mybibfile}
\end{document}